\pdfoutput=1

\documentclass[letterpaper, 10pt, conference]{ieeeconf}  %

\IEEEoverridecommandlockouts                              %

\let\labelindent\relax                                    %

\usepackage{graphicx}     %
\usepackage{amsmath}      %
\usepackage{amssymb}      %
\usepackage{amsfonts}     %
\usepackage{mathtools}    %
\usepackage{enumitem}     %
\usepackage{siunitx}      %
\usepackage{tensor}       %
\usepackage{url}          %
\usepackage{xcolor}       %
\usepackage{xspace}       %
\usepackage[T1]{fontenc}  %
\usepackage{hyperref}     %

\hypersetup{hidelinks, bookmarksnumbered, colorlinks, linkcolor={red!50!black}, citecolor={blue!50!black}, urlcolor={blue!80!black}}

\makeatletter
\let\MYcaption\@makecaption
\makeatother
\usepackage[font=footnotesize]{subcaption}
\makeatletter
\let\@makecaption\MYcaption
\makeatother

\graphicspath{{images/}}
\DeclareGraphicsExtensions{.pdf,.jpeg,.png}

\DeclareMathOperator{\tr}{tr}
\DeclareMathOperator{\sgn}{sgn}
\DeclareMathOperator{\sign}{sign}
\DeclareMathOperator{\asin}{asin}
\DeclareMathOperator{\acos}{acos}
\DeclareMathOperator{\wrap}{wrap}

\DeclareMathOperator{\atantwo}{atan2}
\providecommand{\abs}[1]{\lvert#1\rvert}
\providecommand{\norm}[1]{\lVert#1\rVert}

\newenvironment{lbraced}{\left\lbrace\begin{aligned}}{\end{aligned}\right.}

\newcommand{\vect}[1]{\mathbf{#1}} %
\newcommand{\vecs}[2]{\tensor*{\vect{#1}}{_{#2}}} %
\newcommand{\vecbs}[3]{\tensor*[^{#1}]{\vect{#2}}{_{#3}}} %
\newcommand{\compbs}[3]{\tensor*[^{#1}]{#2}{_{#3}}} %
\newcommand{\rotb}[3]{\tensor*[^{#1}_{#2}]{#3}{}} %
\newcommand{\fr}[1]{\{#1\}\xspace} %
\newcommand{\conj}{^{*}} %
\newcommand{\trans}{^T\!} %
\newcommand{\F}{\mathbb{F}} %
\newcommand{\I}{\mathbb{I}} %
\newcommand{\Q}{\mathbb{Q}} %
\newcommand{\R}{\mathbb{R}} %
\newcommand{\T}{\mathbb{T}} %
\newcommand{\st}{:} %
\newcommand{\vhat}{\vect{\hat{v}}} %
\newcommand{\half}{\tfrac{1}{2}} %
\newcommand{\hpi}{\tfrac{\pi}{2}} %
\newcommand{\defeq}{\equiv} %
\newcommand{\degreem}{^{\circ}} %

\newcommand{\seclabel}[1]{\label{sec:#1}}

\newcommand{\figlabel}[1]{\label{fig:#1}}

\newcommand{\eqnlabel}[1]{\label{eqn:#1}}
\newcommand{\enumlabel}[1]{\label{enum:#1}}
\newcommand{\secref}[1]{Section~\ref{sec:#1}\xspace}

\newcommand{\figref}[1]{Fig.~\ref{fig:#1}\xspace}

\newcommand{\eqnref}[1]{(\ref{eqn:#1})\xspace}
\newcommand{\enumref}[1]{\ref{enum:#1}\xspace}

\newcommand{\eqnrefs}[2]{(\ref{eqn:#1}--\ref{eqn:#2})\xspace}

\newcommand{\nop}{NimbRo\protect\nobreakdash-OP\xspace}

\newcommand{\term}[1]{\emph{#1}\xspace}
\newcommand{\degree}{$\degreem$\xspace}

\title{\LARGE \bf Fused Angles: A Representation of Body Orientation for Balance}

\author{Philipp Allgeuer and Sven Behnke%
\thanks{All authors are with the Autonomous Intelligent Systems (AIS) Group, Computer Science Institute VI,
        University of Bonn, Germany. Email: {\tt\small pallgeuer@ais.uni-bonn.de}. This work was partially
        funded by grant BE 2556/10 of the German Research Foundation (DFG).}}

\usepackage{eso-pic}

\AtBeginDocument{\AddToShipoutPictureFG*{\AtTextUpperLeft{\put(0,\LenToUnit{9pt}){\parbox{\textwidth}{\centering\bfseries
International Conference on Intelligent Robots and Systems (IROS), Hamburg, Germany, 2015
}}}}}
\begin{document}

\maketitle
\thispagestyle{empty}
\pagestyle{empty}

\begin{abstract}
The parameterisation of rotations in three dimensional Euclidean space is an 
area of applied mathematics that has long been studied, dating back to the 
original works of Euler in the 18$^{\text{th}}\mspace{-3mu}$ century. 
As such, many ways of parameterising a rotation have been developed over the 
years. Motivated by the task of representing the orientation of a balancing 
body, the fused angles parameterisation is developed and introduced in this 
paper. This novel representation is carefully defined both mathematically and 
geometrically, and thoroughly investigated in terms of the properties it 
possesses, and how it relates to other existing representations. A second 
intermediate representation, tilt angles, is also introduced as a natural 
consequence thereof.
\end{abstract}

\vspace{-0.3em}
\section{Introduction}
\seclabel{introduction}

Numerous ways of representing a rotation in three-dimensional Euclidean space 
have been developed and refined over the years. Many of these representations, 
also referred to as parameterisations, arose naturally from classical 
mathematics and have found widespread use in areas such as physics, engineering 
and robotics. Prominent examples of such representations include rotation 
matrices, quaternions and Euler angles. In this paper, a new parameterisation of 
the manifold of all three-dimensional rotations is proposed. This 
parameterisation, referred to as \term{fused angles}, was motivated by the 
analysis and control of the balance of bodies in 3D, and the shortcomings of the 
various existing rotation representations to describe the state of balance in an 
intuitive and problem-relevant way. More specifically, the advent of fused 
angles was to address the problem of representing the orientation of a body in 
an environment where there is a clear notion of what is `up', defined 
implicitly, for example, through the presence of gravity. An orientation is just 
a rotation relative to some global fixed frame however, so fused angles can 
equally be used to represent any arbitrary three-dimensional rotation, much like 
Euler angles, for instance, can be used for both purposes. The shortcomings of 
Euler angles, however, that make them unsuitable for this balance-inspired task
are discussed in detail in \secref{review_eulerangles}.

When analysing the balance state of a body, such as for example of a humanoid 
robot, it is very helpful to be able to work with a parameterisation of the 
orientation that yields information about the components of the rotation within 
each of the three major planes, i.e.\ within the $\vect{x}\vect{y}$, 
$\vect{y}\vect{z}$ and $\vect{x}\vect{z}$ planes (see \figref{fused_teaser}). 
These components of rotation can be conceptually thought of as a way of 
simultaneously quantifying the `amount of rotation' about the individual axes. 
It is desirable for these components to each offer a useful geometric 
interpretation, and behave intuitively throughout the rotation space, most 
critically not sacrificing axisymmetry within the horizontal $\vect{x}\vect{y}$ 
plane by the introduction of a clear sequential order of rotations. The notion 
of fusing individual rotation components in a way that avoids such an order 
motivated the term `fused angles'. Quaternions, a common choice of 
parameterisation in computational environments, clearly do not address these 
requirements, as elucidated in \secref{review_quat}.

\begin{figure}[!tb]
\vspace{-0.4em}
\parbox{\linewidth}{\centering\includegraphics[width=1.00\linewidth]{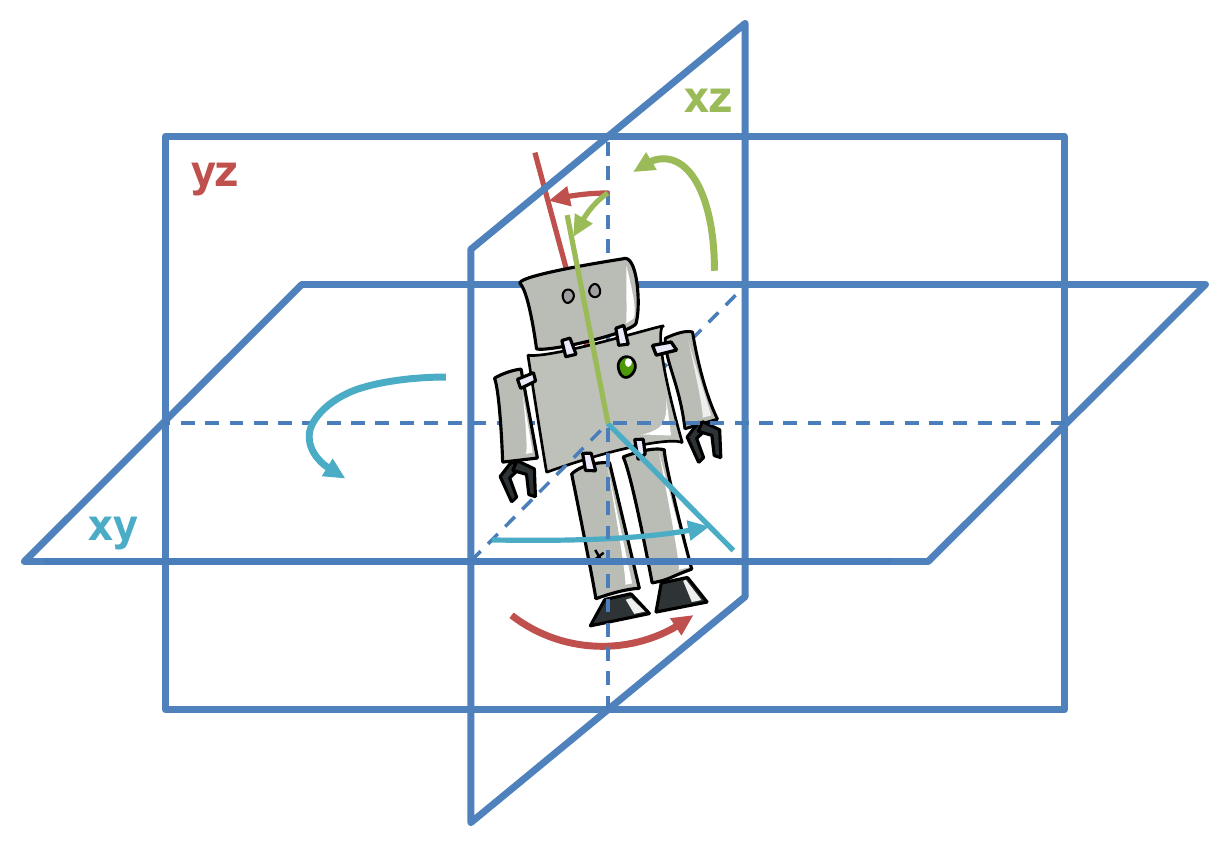}}
\caption{Fused angles are a way of decomposing a rotation into three 
concurrently acting components, in such a way that it gives insight into how 
rotated a body is in each of the three major planes.}
\figlabel{fused_teaser}
\vspace{-0.8em}
\end{figure}

The fused angles rotation representation has to date found a number of uses. 
Most recently in work published by the same authors, an attitude estimator was 
formulated that internally relied on the concept of fused angles 
\cite{Allgeuer2014}. The open source ROS software for the \nop humanoid robot 
\cite{Allgeuer2013a}, developed by the University of Bonn, also relies on the 
use of fused angles, most notably in the areas of state estimation and walking. 
Furthermore, a Matlab and Octave library \cite{MatOctRotLibGithub} targeted at 
the numerical and computational handling of all manners of three-dimensional 
rotation representations, including fused angles, has been released.\footnote{ 
{\scriptsize\url{https://github.com/AIS-Bonn/matlab_octave_rotations_lib}}\\Also 
C++ Library: {\scriptsize\url{https://github.com/AIS-Bonn/rot_conv_lib}}} This 
library is intended to serve as a common reference for the implementation in 
other programming languages of a wide range of conversion and computation 
functions. It is seen by the authors as a test bed to support the development of 
new rotation-related algorithms.

The convention that the global z-axis points in the `up' direction relative to 
the environment is used in this paper. As mentioned previously, this accepted 
`up' direction will almost always be defined as the antipodal direction 
of gravity. This ensures that definitions such as that of \term{fused yaw} make 
terminological sense in consideration of the true rotation of a body relative to 
its environment. All derived formulas and results could easily be rewritten 
using an alternative convention if this were to be desired.

The contribution of this paper lies in the introduction of the novel concept of 
fused angles for the representation of rotations. A further contribution is the 
concept of tilt angles (see \secref{tilt_angles_geometric}), an intermediary 
representation that emerges naturally from the derivation of the former.

\section{Review of Existing Rotation Representations}
\seclabel{review}

Many ways of representing 3D rotations in terms of a finite set of parameters
exist. Different representations have different advantages and disadvantages,
and which representation is suitable for a particular application depends on a
wide range of considerations. Such considerations include:
\begin{itemize}
 \item Ease of geometric interpretation, in particular in a form that is
       relevant to the particular problem,
 \item The range of singularity-free behaviour,
 \item Computational efficiency in terms of common operations such as rotation
       composition and vector rotation,
 \item Mathematical convenience, in terms of numeric and algebraic complexity
       and manipulability, and
 \item Algorithmic convenience, in the sense of a representation potentially
       possessing properties that can conveniently be exploited for a particular
       algorithm.
\end{itemize}
A wide range of existing rotation representations are reviewed in this section 
as a basis for comparison. Due to the dimensionality of the space of 3D 
rotations, a minimum of three parameters is required for any such 
representation. A representation with exactly three parameters is referred to as 
\term{minimal}, while other representations with a greater number of parameters 
are referred to as \term{redundant}.

\subsection{Rotation Matrices}
\seclabel{review_rotmat}

A rotation can be represented as a linear transformation of coordinate frame 
basis vectors, expressed in the form of an orthogonal matrix of unit 
determinant. Due to the strong link between such transformation matrices and the 
theory of direction cosines, the name Direction Cosine Matrix is also sometimes 
used. The space of all rotation matrices is called the special orthogonal group 
$\text{SO}(3)$, and is defined as
\begin{equation}
\text{SO}(3) = \{R \in \R^{3\times3} \st R\trans R=\I, \,\det(R)=1\}.
\end{equation}
Rotation of a vector $\vect{v} \in \R^3$ by a rotation matrix is given by matrix
multiplication. For a rotation from coordinate frame \fr{G} to \fr{B}, we have that
\begin{equation}
\rotb{G}{B}{R} =
\begin{bmatrix}
\vecbs{G}{x}{B} & \mspace{-5mu}\vecbs{G}{y}{B} & \mspace{-5mu}\vecbs{G}{z}{B}
\end{bmatrix}
 = 
\begin{bmatrix}
\vecbs{B}{x}{G} & \mspace{-5mu}\vecbs{B}{y}{G} & \mspace{-5mu}\vecbs{B}{z}{G}
\end{bmatrix}\trans\!,
\eqnlabel{rotmatrowcol}
\end{equation}
where $\vecbs{G}{y}{B}$, for example, is the column vector corresponding to the
y-axis of frame \fr{B}, expressed in the coordinates of frame \fr{G}. The
notation $\rotb{G}{B}{R}$ refers to the relative rotation from \fr{G} to \fr{B}.
With nine parameters, rotation matrices are clearly a redundant parameterisation
of the rotation space. They are quite useful in that they are free of
singularities and trivially expose the basis vectors of the fixed and rotated
frames, but for many tasks they are not as computationally and numerically
suitable as other representations.

\subsection{Axis-Angle and Rotation Vector Representations}
\seclabel{review_axisangle_rotvec}

By Euler's rotation theorem \cite{Palais2009}, every rotation in the 
three-dimensional Euclidean space $\R^3$ can be expressed as a single rotation 
about some axis. As such, each rotation can be mapped to a pair 
$(\vect{\hat{u}},\theta) \in S^2\times\R$, where $\vect{\hat{u}}$ is a unit 
vector corresponding to the axis of rotation, and $\theta$ is the magnitude of 
the rotation. Note that $S^2 = \{\vect{v} \in \R^3 \st \norm{\vect{v}}=1\}$, the 
2-sphere, is the set of all unit vectors in $\R^3$. A closely related concept is 
that of the rotation vector, given by $\vect{u} = \theta\vect{\hat{u}}$, which 
encodes the angle of rotation as the magnitude of the vector defining the 
rotation axis. Both the axis-angle and rotation vector representations suffer 
from a general impracticality of mathematical and numerical manipulation. For 
example, no formula for rotation composition exists that is more direct than 
converting to quaternions and back. The Simultaneous Orthogonal Rotations Angle 
(SORA) vector, a slight reformulation of the rotation vector concept in terms of 
virtual angular velocities and virtual time, was presented by Toma\v{z}i\v{c} and 
Stan\v{c}in in \cite{Tomazic2011}. This formulation suffers from drawbacks 
similar to those of the rotation vector representation, which includes a 
discontinuity at rotations of 180\degree, and a general lack of geometric 
intuitiveness.

\subsection{Quaternions}
\seclabel{review_quat}

The set of all quaternions $\mathbb{H}$, and the subset $\Q$ thereof of all
quaternions that represent pure rotations, are defined as
\begin{equation}
\begin{split}
\mathbb{H} &= \{q = (q_0,\vect{q}) \defeq (w,x,y,z) \in \R^4\}, \\
\Q &= \{q \in \mathbb{H} \st \norm{q} = 1\}.
\end{split}
\end{equation}
Quaternion rotations can be related to the axis-angle representation, and
thereby visualised to some degree, using
\begin{equation}
q = (q_0,\vect{q}) = \bigl(\cos{\tfrac{\theta}{2}}, \vect{\hat{u}}\sin{\tfrac{\theta}{2}}\bigr) \in \Q, \eqnlabel{quat_axisangle}
\end{equation}
where $(\vect{\hat{u}},\theta) \in S^2\times\R$ is any axis-angle rotation pair,
and $q$ is the equivalent quaternion rotation. The use of quaternions to express
rotations generally allows for very computationally efficient calculations, and
is grounded by the well-established field of quaternion mathematics. A crucial
advantage of the quaternion representation is that it is free of singularities.
On the other hand however, it is not a one-to-one mapping of the special
orthogonal group, as $q$ and $-q$ both correspond to the same rotation. The
redundancy of the parameters also means that the unit magnitude constraint has
to explicitly and sometimes non-trivially be enforced in numerical computations.
Furthermore, no clear geometric interpretation of quaternions exists beyond the
implicit relation to the axis-angle representation given in
\eqnref{quat_axisangle}. For applications related to the balance of a body,
where questions arise such as `how rotated' a body is in total or within a
particular major plane, the quaternion representation yields no direct insight.

\subsection{Euler Angles}
\seclabel{review_eulerangles}

A step in the right direction of understanding the different components of a 
rotation is the notion of Euler angles. In this representation, the total 
rotation is split into three individual elemental rotations, each about a 
particular coordinate frame axis. The three Euler angles
$(\alpha, \beta, \gamma)$ describing a rotation are the successive magnitudes of 
these three elemental rotations. Many conventions of Euler angles exist, 
depending on the order in which the elemental axis rotations are chosen and 
whether the elemental rotations are taken to be intrinsic (about the rotating 
coordinate frame) or extrinsic (about the fixed coordinate frame). Extrinsic 
Euler angles can easily be mapped to their equivalent intrinsic Euler angles 
representations, and so the two types do not exhibit fundamentally different 
behaviour. If all three coordinate axes are used in the elemental rotations, the 
representation is alternatively known as Tait-Bryan angles, and the three 
parameters are referred to as yaw, pitch and roll, respectively. Tait-Bryan 
angles, although promising at first sight, do not suffice for the representation 
of the orientation of a body in balance-related scenarios. The main reasons for 
this are:
\begin{itemize}
 \item The proximity of the gimbal lock singularity to normal working ranges,
       leading to unwanted artefacts due to increased local parameter
       sensitivity in a widened neighbourhood of the singularity,
 \item The fundamental requirement of an order of elemental rotations, leading
       to asymmetrical definitions of pitch and roll that do not correspond
       in behaviour, and
 \item The asymmetry introduced by the use of a yaw definition that depends on
       the projection of one of the coordinate axes onto a fixed plane, leading
       to unintuitive non-axisymmetric behaviour of the yaw angle.
\end{itemize}
The first listed point is a problem in real life, if for example a bipedal
robot falls down, and thereby comes near the Euler angle singularity.
As an example of the last of the listed points, consider the intrinsic ZYX Euler 
angles representation, recalling that the global 
z-axis points `upwards' (see \secref{introduction}). Consider a body in space, 
assumed to be in its identity orientation, and some arbitrary rotation of the 
body relative to its environment. It would be natural and intuitive to expect 
that the yaw of this relative rotation is independent of the chosen definition 
of the global x and y-axes. This is because the true rotation of the body is 
always the same, regardless of the essentially arbitrary choice of the global x 
and y-axes, and one would expect a well-defined yaw to be a property of the 
rotation, not the axis convention. This is not the case for ZYX Euler yaw 
however, as can be verified by counterexample with virtually any non-degenerate 
case. The yaw component of the fused angles representation, defined in 
\secref{tilt_angles_geometric}, can be proven to satisfy this property.

\subsection{Vectorial Parameterisations}
\seclabel{review_vectorial}

Parameterisations are sometimes developed specifically to exhibit certain 
properties that can be exploited to increase the efficiency of an algorithm. A 
class of such generally more mathematical and abstract rotation representations 
is the family of vectorial parameterisations. Named examples of these include 
the Gibbs-Rodrigues parameters \cite{Bauchau2003} and the Wiener-Milenkovi\'c 
parameters \cite{Trainelli2004}, also known as the conformal rotation vector 
(CRV). Such parameterisations derive from mathematical identities such as the 
Euler-Rodrigues formula \cite{Argyris1982}, and as such do not in general have 
any useful geometric interpretation, and find practical use in only very 
specific applications. Detailed derivations and analyses of vectorial 
parameterisations can be found in \cite{Bauchau2003} and \cite{Trainelli2004}.

\section{Fused Angles}
\seclabel{fused_angles_defn}

Fused angles were motivated by the lack of an existing 3D rotation formalism 
that naturally deals with the dissolution of a complete rotation into parameters 
that are specifically and geometrically relevant to the balance of a body, and 
that does not introduce order-based asymmetry in the parameters. None of the 
representations discussed in \secref{review} satisfy this property. The unwanted 
artefacts in the existing notions of yaw (see \secref{review_eulerangles}) also 
led to the need for a more suitable, stable and axisymmetric definition of yaw.

\subsection{Geometric Definition of Tilt Angles}
\seclabel{tilt_angles_geometric}

\begin{figure}[!tb]
\parbox{\linewidth}{\centering\includegraphics[width=0.95\linewidth]{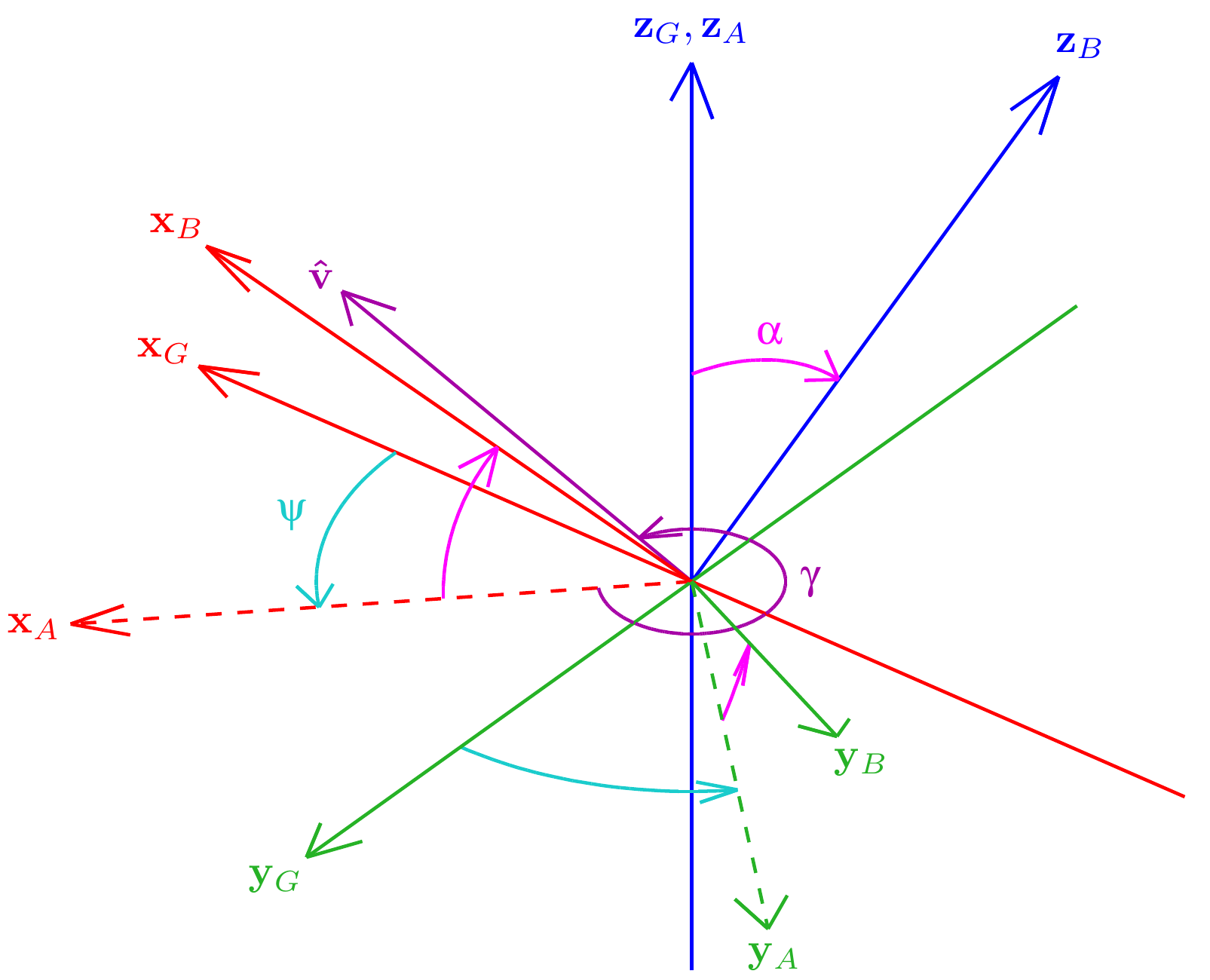}}
\caption{Definition of the tilt rotation and tilt angles parameters of the 
rotation from \fr{G} to \fr{B}. $\psi$ is the fused yaw, $\gamma$ is the tilt 
axis angle, $\alpha$ is the tilt angle and $\vhat$ is the tilt axis. The 
intermediate frame \fr{A} is constructed by rotating \fr{B} such that 
$\vecs{z}{B}$ rotates directly onto $\vecs{z}{G}$.}
\figlabel{tilt_angles_frames}
\vspace{-0.3em}
\end{figure}

We begin by defining an intermediate rotation representation, referred to as 
\term{tilt angles}. The tilt angles parameter definitions are illustrated in 
\figref{tilt_angles_frames}. Note that we follow the convention that, for 
example,
\begin{equation}
\vecbs{G}{z}{B} = \bigl(\compbs{G}{z}{Bx},\compbs{G}{z}{By},\compbs{G}{z}{Bz}\bigr)
\end{equation}
denotes the unit vector corresponding to the positive z-axis of a frame \fr{B}, 
expressed in the coordinates of a frame \fr{G}. The absence of a coordinate 
basis qualifier, such as for example in the notation `$\vecs{z}{B}$', implies 
that a vector is by default expressed relative to the global fixed frame.

Let \fr{G} denote the global fixed frame, defined with the convention that the 
global z-axis points upwards in the environment, as discussed in 
\secref{introduction}. We define \fr{B} to be the body-fixed coordinate frame. 
For an identity orientation of the body, the frames \fr{G} and \fr{B} should 
clearly coincide.

As $\vecs{z}{G}$ and $\vecs{z}{B}$ are vectors in $\R^3$, a rotation about an 
axis perpendicular to both vectors exists that maps $\vecs{z}{G}$ onto 
$\vecs{z}{B}$. Note that this is a different condition to mapping \fr{G} onto \fr{B}.
We choose an axis-angle representation $(\vhat,\alpha)$ (see 
\secref{review_axisangle_rotvec}) of this \term{tilt rotation} such that
$\alpha \in [0,\pi]$. The angle $\alpha$ is referred to as the \term{tilt angle} 
of \fr{B}, and the vector $\vhat$ is referred to as the \term{tilt axis} of 
\fr{B}. We define coordinate frame \fr{A} to be the frame that results when we 
apply the inverse of the tilt rotation to \fr{B}. By definition
$\vecs{z}{A} = \vecs{z}{G}$, so it follows that $\vhat$---and trivially also 
$\vecs{x}{A}$---must lie in the $\vecs{x}{A}\vecs{y}{A}$ plane. The angle 
$\gamma$ about $\vecs{z}{A}$ from $\vecs{x}{A}$ to $\vhat$ (see 
\figref{tilt_angles_frames}) is referred to as the \term{tilt axis angle} of 
\fr{B}. It is easy to see that the tilt rotation from \fr{A} to \fr{B} is 
completely defined by the parameter pair $(\gamma,\alpha)$.

We now note that the rotation from \fr{G} to \fr{A} is one of pure yaw, that is, 
a pure z-rotation, and so define the angle $\psi$ about $\vecs{z}{G}$ from 
$\vecs{x}{G}$ to $\vecs{x}{A}$ (see \figref{tilt_angles_frames}) as the 
\term{fused yaw} of \fr{B}. It is important to note that the choice of using the 
x-axes in this definition of yaw is arbitrary, and a similar definition using 
the y-axes would be completely equivalent. The complete tilt angles 
representation of the rotation from \fr{G} to \fr{B} is now defined as
\begin{equation}
\rotb{G}{B}{T} = (\psi,\gamma,\alpha) \in (-\pi,\pi] \times (-\pi,\pi] \times [0,\pi] \defeq \T. \eqnlabel{tiltdefn}
\end{equation}
The identity tilt angles rotation is given by $(0,0,0) \in \T$.

It can be seen from the method of construction that all rotations possess a tilt 
angles representation, although it is not always necessarily unique. Most 
notably, when $\alpha = 0$, the $\gamma$ parameter can be arbitrary with no 
effect.

\subsection{Geometric Definition of Fused Angles}
\seclabel{fused_angles_geometric}

\begin{figure}[!tb]
\parbox{\linewidth}{\centering\includegraphics[width=0.95\linewidth]{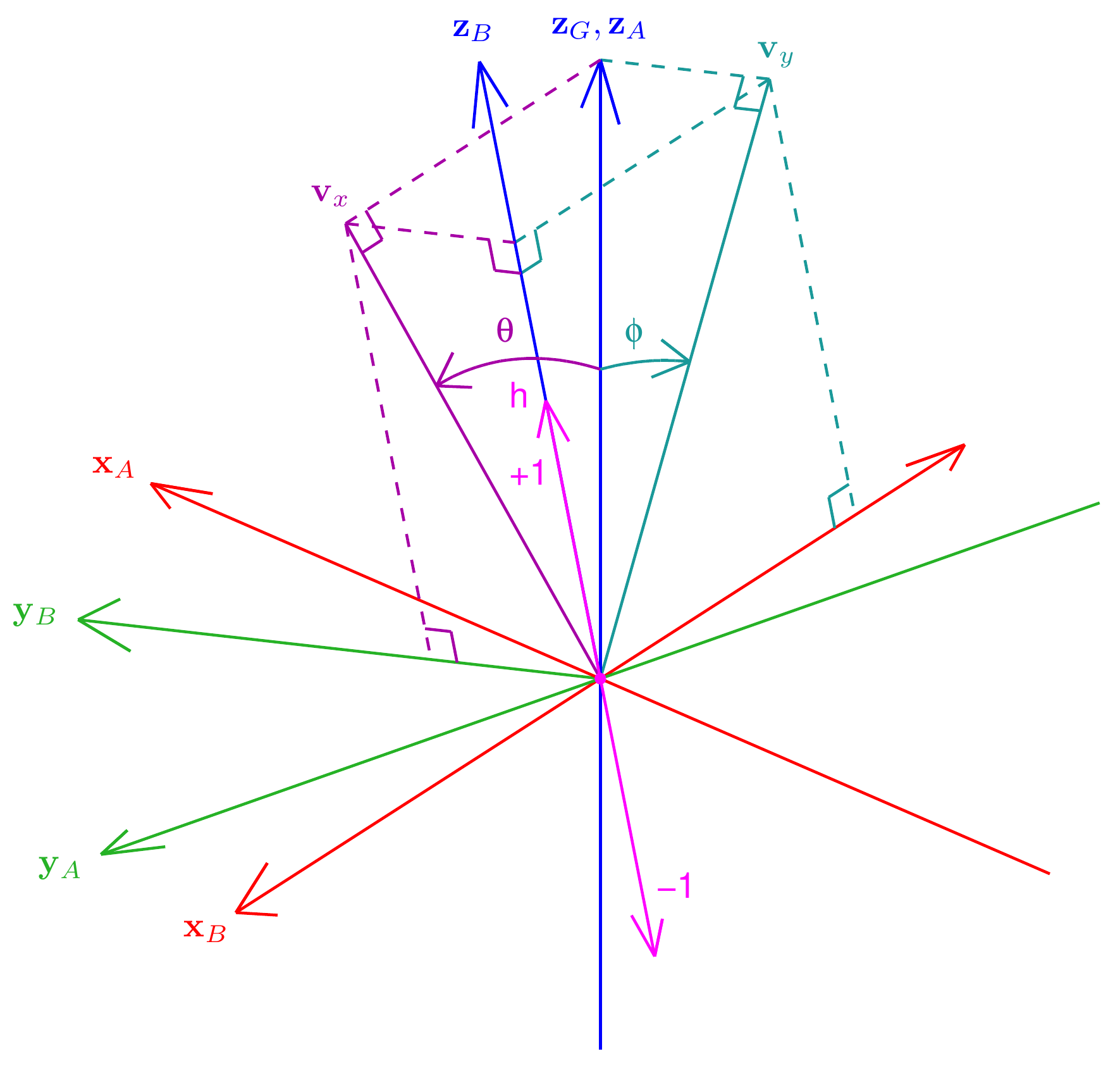}}
\caption{Definition of the fused angles parameters $(\theta,\phi,h)$ that 
describe the tilt rotation component of the rotation from \fr{G} to \fr{B}. 
\fr{A} is the same intermediate frame as from the geometric definition of tilt 
angles (see \figref{tilt_angles_frames}). The $\vecs{x}{G}$ and $\vecs{y}{G}$ 
axes are not shown in the figure for visual simplicity. $\theta$ is the fused 
pitch, $\phi$ is the fused roll, $h$ is the hemisphere, $\vecs{v}{x}$ is the 
projection of $\vecs{z}{G}$ onto the $\vecs{y}{B}\vecs{z}{B}$ plane, and 
$\vecs{v}{y}$ is the projection of $\vecs{z}{G}$ onto the 
$\vecs{x}{B}\vecs{z}{B}$ plane. Geometrically it can be seen that 
$(\theta,\phi,h)$ depends only on the direction of $\vecs{z}{G}$ relative to 
\fr{B}, that is, $\vecbs{B}{z}{G}$.}
\figlabel{fused_angles_frames}
\vspace{-0.3em}
\end{figure}

To remedy the possible ambiguity in the tilt angles parameters and work towards 
a more robust rotation representation, we introduce the concepts of \term{fused 
pitch} and \term{fused roll}. For reference, the relevant fused angles parameter 
definitions are illustrated in \figref{fused_angles_frames}.

Let $\vecs{v}{x}$ and $\vecs{v}{y}$ be the projections of the $\vecs{z}{G}$ 
vector onto the body-fixed $\vecs{y}{B}\vecs{z}{B}$ and $\vecs{x}{B}\vecs{z}{B}$ 
planes respectively. We define the fused pitch of \fr{B} as the angle $\theta$ 
between $\vecs{z}{G}$ and $\vecs{v}{x}$, of sign 
$-\!\sgn\bigl(\vecbs{B}{z}{Gx}\bigr)$. By logical completion, the magnitude of 
$\theta$ is taken to be $\hpi$ if $\vecs{v}{x} = \vect{0}$. We similarly define 
the fused roll of \fr{B} as the angle $\phi$ between $\vecs{z}{G}$ and 
$\vecs{v}{y}$, of sign $\sgn\bigl(\vecbs{B}{z}{Gy}\bigr)$. The magnitude of 
$\phi$ is taken to be $\hpi$ if $\vecs{v}{y} = \vect{0}$. Conceptually, fused 
pitch and roll can be thought of simply as the angles between $\vecs{z}{G}$ and 
the $\vecs{y}{B}\vecs{z}{B}$ and $\vecs{x}{B}\vecs{z}{B}$ planes respectively. 
Note that this definition of fused pitch and roll is invariant to the entire 
body-fixed frame \fr{B} being yawed, as one would expect.

From inspection of the geometric definitions, it can be seen that the fused pitch 
and roll only uniquely specify a tilt rotation up to the z-hemisphere, that is, 
whether $\vecs{z}{B}$ and $\vecs{z}{G}$ are mutually in the same hemisphere or 
not. To resolve this ambiguity, the \term{hemisphere} of a rotation (see 
\figref{fused_angles_frames}) is defined as
$\sign\bigl(\vecbs{B}{z}{Gz}\bigr) \defeq \sign\bigl(\vecbs{G}{z}{Bz}\bigr)$, 
where we define
\begin{equation}
\sign(x) = 
\begin{cases}
1 & \text{if $x \geq 0$}, \\
0 & \text{if $x < 0$}.
\end{cases}
\end{equation}
Note that $\sign$ differs to the normal definition of a sign function in that 
$\sign(0) = 1$, whereas $\sgn(0) = 0$. This modified sign function is used 
throughout the remainder of this paper wherever clear distinction from the 
normal sign function is required.

Using the concept of the hemisphere of a rotation, $(\theta,\phi,h)$ becomes a 
complete description of the tilt rotation component of a rotation. As such, 
together with the fused yaw $\psi$, the complete fused angles representation of 
the rotation from \fr{G} to \fr{B} can now be defined as
\begin{align}
\rotb{G}{B}{F} &= (\psi,\theta,\phi,h) \notag\\
&\in (-\pi,\pi] \times [-\hpi,\hpi] \times [-\hpi,\hpi] \times \{-1,1\} \defeq \hat{\F}. \eqnlabel{fuseddefn}
\end{align}
The identity fused angles rotation is given by $(0,0,0,1) \in \hat{\F}$. The 
$(\theta,\phi,h)$ triplet in \eqnref{fuseddefn} replaces the $(\gamma,\alpha)$ 
pair in \eqnref{tiltdefn} to define the tilt rotation component of a general 
rotation.

It can be observed from the geometric definitions above that the tilt rotation 
depends only on the direction of $\vecs{z}{G}$ relative to frame \fr{B}---that 
is, only on $\vecbs{B}{z}{G}$. This, for example, means that the bottom row of the 
rotation matrix $\rotb{G}{B}{R}$ (representing the rotation from \fr{G} to 
\fr{B}) can be completely identified with the tilt rotation component of that 
rotation. Interestingly, it can also be seen that the direction of $\vecs{z}{G}$ 
relative to frame \fr{B} is precisely what an accelerometer attached to the body 
would measure under the assumption of quasi-static conditions. In this way, 
accelerometer measurements can easily be mapped to measurements of 
$(\theta,\phi,h)$ and/or $(\gamma,\alpha)$.

\subsection{Mathematical Definition of Fused Angles and Tilt Angles}
\seclabel{fused_tilt_mathematical}

\begin{figure}[!tb]
\parbox{\linewidth}{\centering\includegraphics[width=0.95\linewidth]{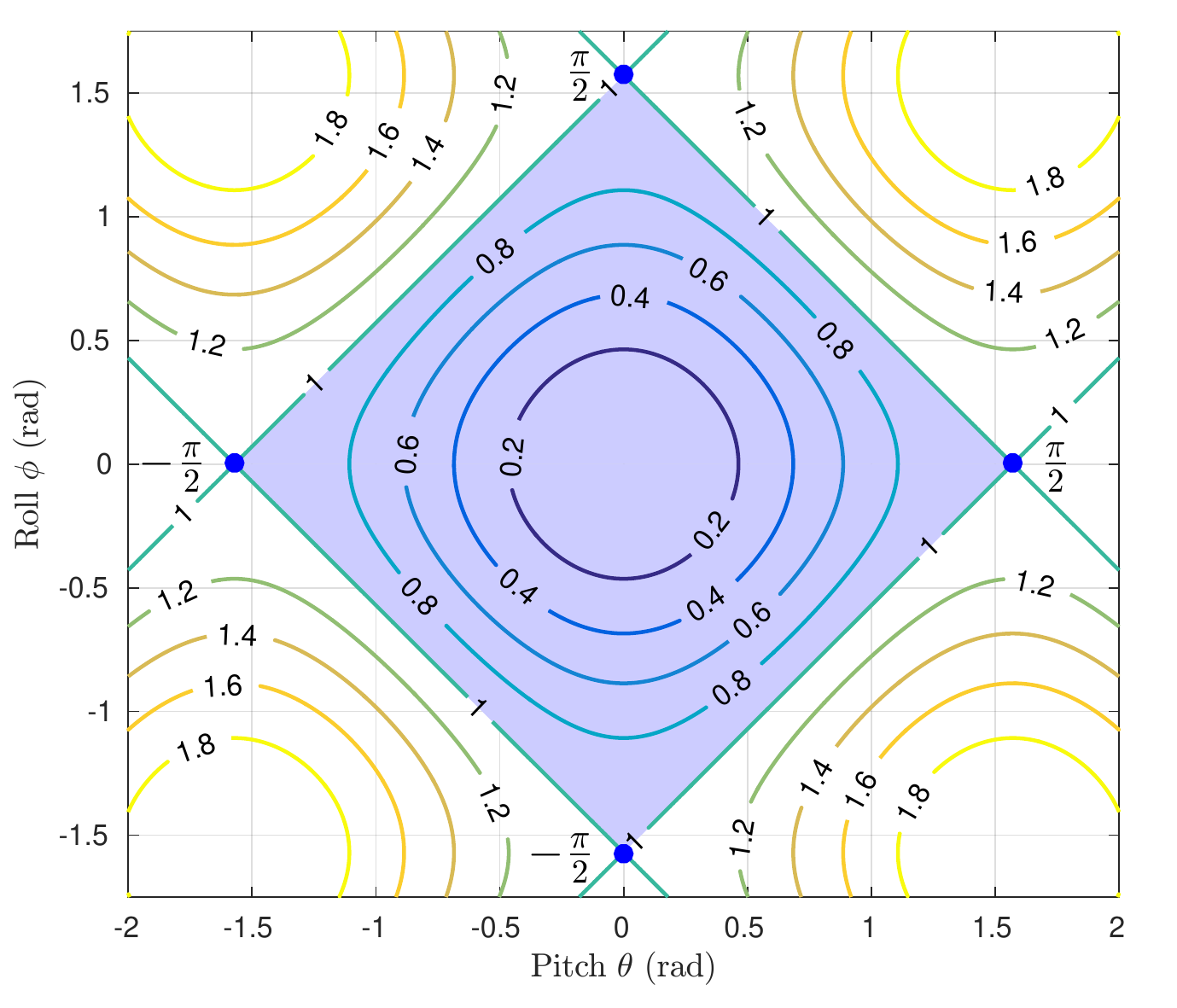}}
\caption{Level sets of the function $f_s(\theta,\phi) = \sin^2\!\theta + \sin^2\!\phi$,
demonstrating that the sine sum criterion $f_s(\theta,\phi) \leq 1$ is 
equivalent to the simpler inequality $\abs{\theta} + \abs{\phi} \leq \hpi$, 
indicated by the shaded region in the plot. The shaded region is the domain of 
$(\theta,\phi)$ for the fused angles representation.}
\figlabel{pitch_roll_domain}
\vspace{-1em}
\end{figure}

Based on the given geometric definitions, the following expressions can be 
derived as an alternative mathematical definition of the tilt angles tilt rotation 
parameters:
\begin{alignat}{2}
\gamma &= \atantwo\bigl(-\compbs{B}{z}{Gx},\compbs{B}{z}{Gy}\bigr) && \in (-\pi,\pi], \eqnlabel{zGtogamma}\\
\alpha &= \acos\bigl(\compbs{B}{z}{Gz}\bigr) && \in [0,\pi]. \eqnlabel{zGtoalpha}
\end{alignat}
Similarly, alternative mathematical definitions for the fused angles tilt rotation 
parameters can be derived to be
\begin{alignat}{2}
\theta &= \asin\bigl(-\compbs{B}{z}{Gx}\bigr) && \in [-\hpi,\hpi], \eqnlabel{zGtotheta}\\
\phi &= \asin\bigl(\compbs{B}{z}{Gy}\bigr) && \in [-\hpi,\hpi], \eqnlabel{zGtophi}\\
h &= \sign\bigl(\compbs{B}{z}{Gz}\bigr) && \in \{-1,1\}. \eqnlabel{zGtohemi}
\end{alignat}
The analysis for the fused yaw parameter is slightly more complex, but with the 
use of cases one can nonetheless mathematically define it as
\begin{equation}
\psi = 
\begin{cases}
\wrap\bigl(\atantwo\bigl(\compbs{G}{z}{Bx},-\compbs{G}{z}{By}\bigr) - \gamma\bigr) & \text{if $\alpha \neq 0$}, \\
\atantwo\bigl(\compbs{G}{x}{By},\compbs{G}{x}{Bx}\bigr) & \text{if $\alpha = 0$},
\end{cases}
\eqnlabel{psidefn}
\end{equation}
where $\wrap$ is a function that wraps an angle to $(-\pi,\pi]$. An alternative 
mathematical definition for fused yaw, namely \eqnref{quattopsitheta}, is 
presented later in \secref{conversions}.

It can be seen from \eqnrefs{zGtotheta}{zGtohemi} and the unit norm condition that 
$\vecbs{B}{z}{G}$ is given by a well-defined multivariate function
$f_z : (\theta,\phi,h) \mapsto \vecbs{B}{z}{G}$, described by
\begin{equation}
\vecbs{B}{z}{G} = \Bigl(-\sin\theta, \sin\phi, h\sqrt{1-\sin^2\theta-\sin^2\phi}\Bigr),
\end{equation}
where for obvious reasons we must have $\sin^2\theta + \sin^2\phi \leq 1$. This 
inequality is referred to by the authors as the \term{sine sum criterion}, and is precisely 
equivalent to
\begin{equation}
\abs{\theta} + \abs{\phi} \leq \hpi.
\end{equation}
Given that by definition $\theta,\phi \in [-\hpi,\hpi]$, this equivalence can be 
seen by plotting the level sets of the multivariate function
\begin{equation}
f_s(\theta,\phi) = \sin^2\theta + \sin^2\phi,
\end{equation}
and finding the region where $f_s(\theta,\phi) \leq 1$. The resulting plot is 
shown in \figref{pitch_roll_domain}. The domain of $f_z$ is the restriction 
of $[-\hpi,\hpi] \times [-\hpi,\hpi] \times \{-1,1\}$ to
$\abs{\theta} + \abs{\phi} \leq \hpi$, and the universal set of all fused 
angles, $\F$, is a similar restriction of $\hat{\F}$---that is, a restriction by 
the sine sum criterion.

\subsection{Visualisation of Fused Angles}
\seclabel{fused_angles_vis}

\begin{figure*}
\centering
\subcaptionbox{Constant fused pitch cone for $\theta = 50\degreem$\figlabel{fused_cones_pitch}}[0.32\linewidth][l]{\includegraphics[width=0.32\linewidth]{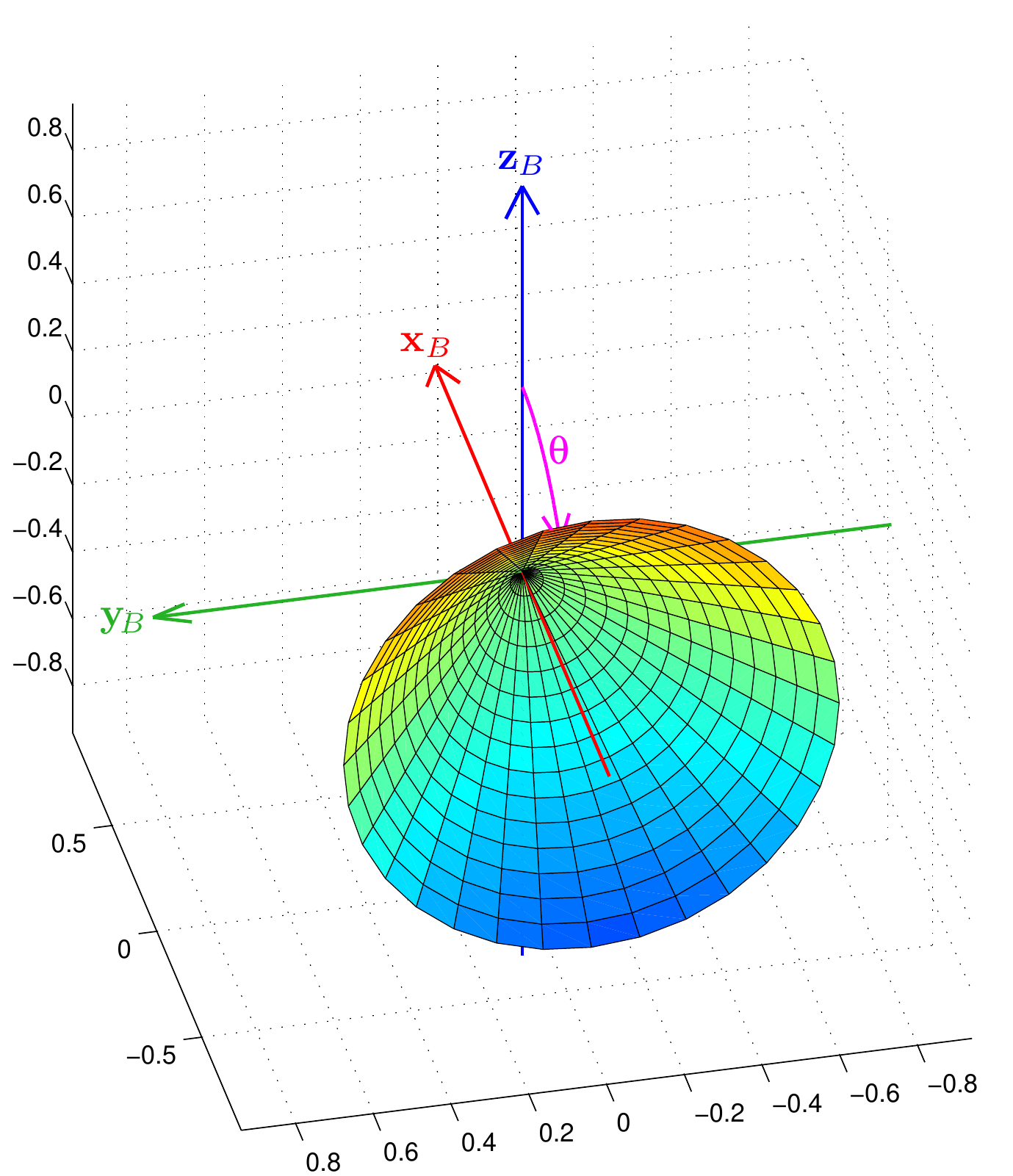}}
\subcaptionbox{Constant fused roll cone for $\phi = 32\degreem$\figlabel{fused_cones_roll}}[0.32\linewidth][c]{\includegraphics[width=0.32\linewidth]{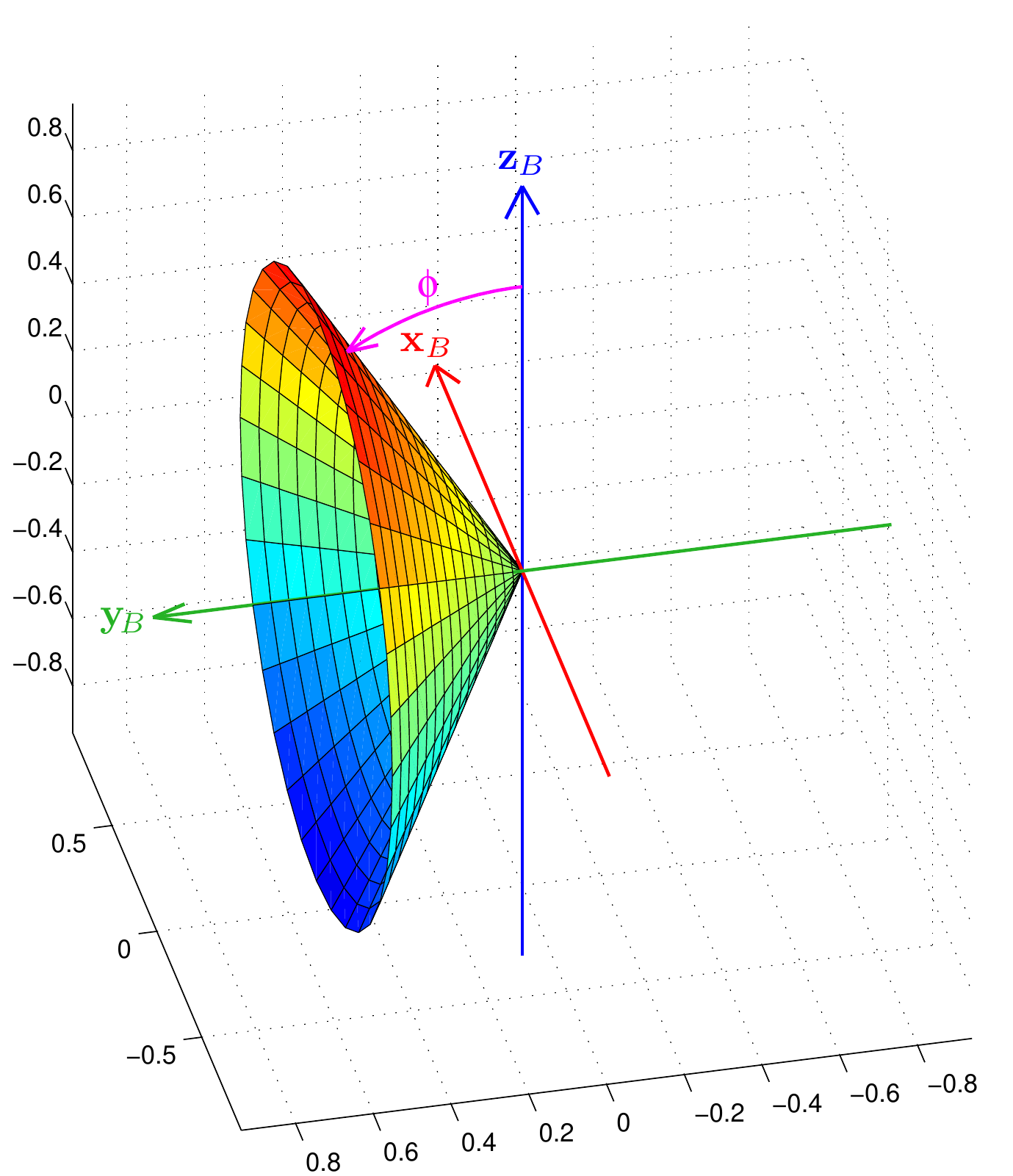}}
\subcaptionbox{Intersection of fused pitch and roll cones\figlabel{fused_cones_pitch_roll}}[0.32\linewidth][r]{\includegraphics[width=0.32\linewidth]{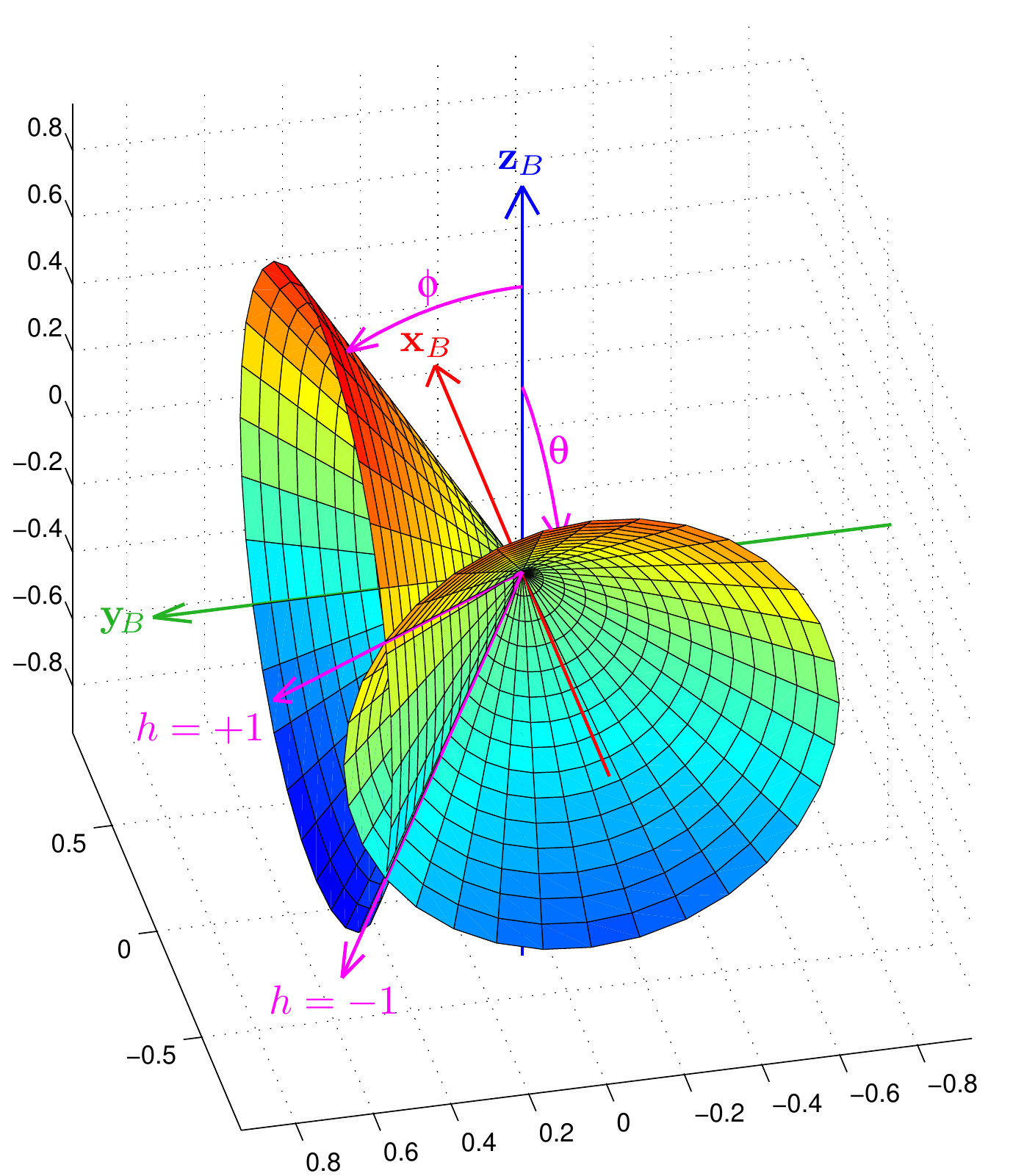}}
\\\medskip
\subcaptionbox{Constant hemisphere locus for $h = +1$\figlabel{fused_cones_hemip}}[0.32\linewidth][l]{\includegraphics[width=0.32\linewidth]{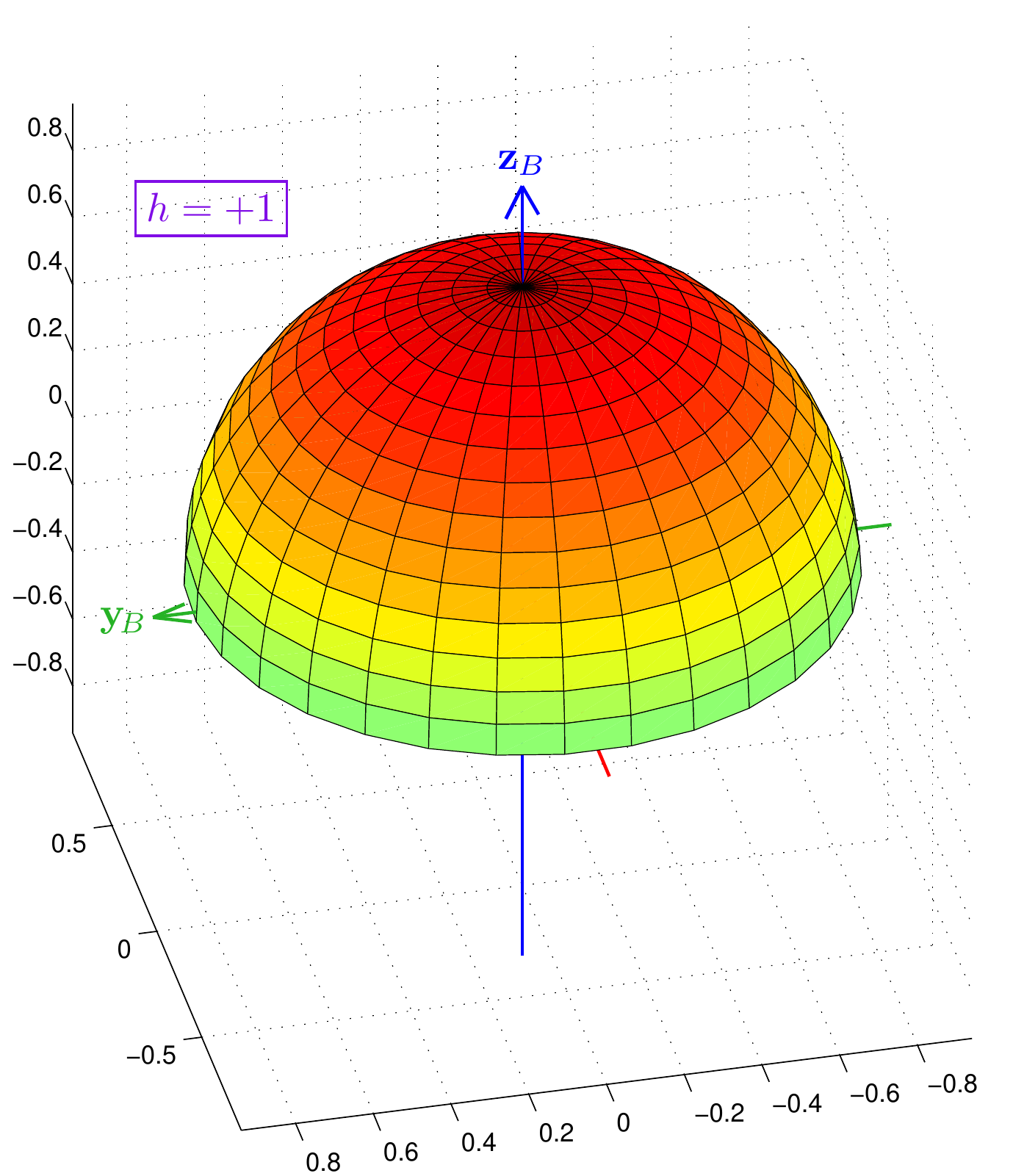}}
\subcaptionbox{Constant hemisphere locus for $h = -1$\figlabel{fused_cones_hemin}}[0.32\linewidth][c]{\includegraphics[width=0.32\linewidth]{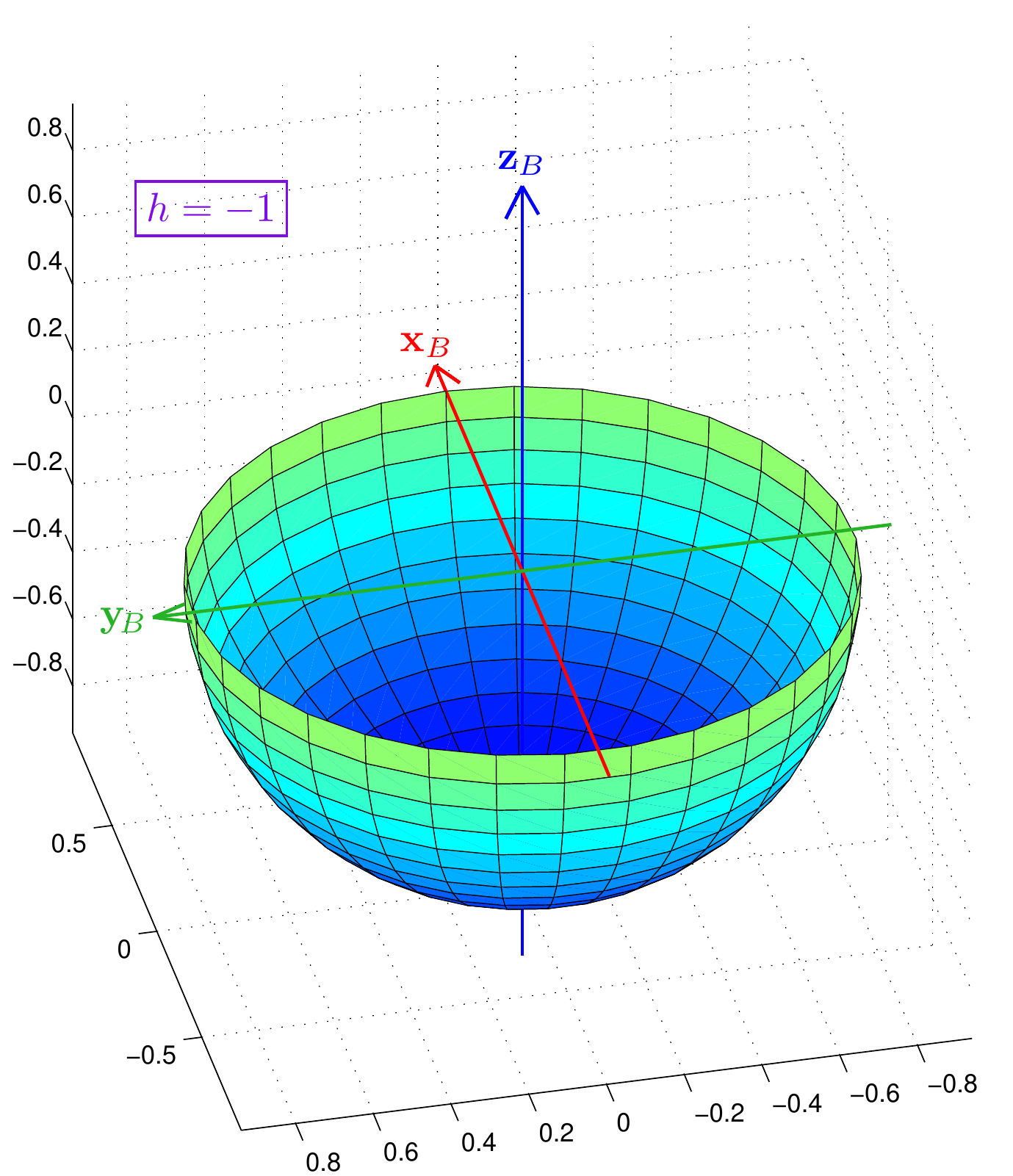}}
\subcaptionbox{Intersection of cones and hemisphere locus\figlabel{fused_cones_pitch_roll_hemi}}[0.32\linewidth][r]{\includegraphics[width=0.32\linewidth]{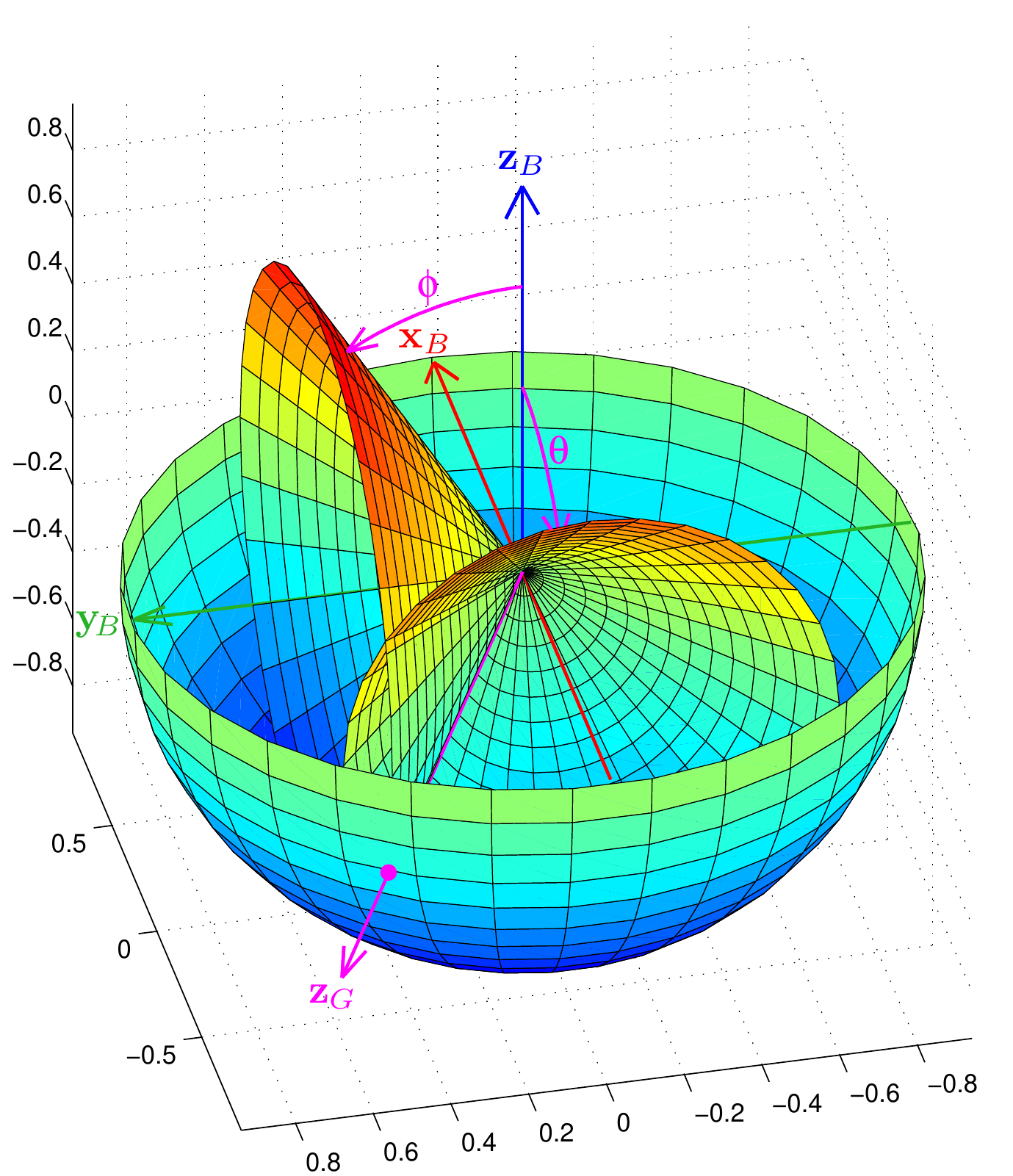}}
\\\smallskip
\caption{Plots of the 3D locus of $\vecbs{B}{z}{G}$ for (a) constant fused pitch 
$\theta$, (b) constant fused roll $\phi$, (d) constant hemisphere $h = +1$, and 
(e) constant hemisphere $h = -1$. (c) shows how loci of constant fused pitch and 
roll intersect at (at most) two points, illustrating how $(\theta,\phi)$ 
specifies $\vecbs{B}{z}{G}$ up to two possible choices. (f) shows how 
intersecting again with the applicable hemisphere locus uniquely resolves 
$\vecbs{B}{z}{G}$. Note that the locus of constant fused pitch in (a) is a cone 
with its opening facing out of the page, much like the opening of the cone of 
constant fused roll in (b) is facing towards the left.}
\figlabel{fused_cones}
\vspace{-0.5em}
\end{figure*}

The fused yaw parameter $\psi$ is best visualised precisely as defined and 
illustrated in \figref{tilt_angles_frames}. The remaining fused angles 
parameters, $(\theta,\phi,h)$, are also well visualised based on their geometric 
definition shown in \figref{fused_angles_frames}, but can alternatively be 
envisioned as loci of $\vecbs{B}{z}{G}$. \figref{fused_cones} shows surface 
plots of the manifolds that are generated by independently taking the image of 
$f_z(\theta,\phi,h)$ for constant fused pitch $\theta$, fused roll $\phi$ and 
hemisphere $h$. The surfaces that result can be seen to be single-ended cones 
and hemispheres. It is important to note that the plots are in the body-fixed 
frame \fr{B}, and not in the global fixed frame \fr{G}. 
\figref{fused_cones_pitch_roll} and \figref{fused_cones_pitch_roll_hemi} show 
how combining specifications of $\theta$, $\phi$ and $h$ acts to resolve a 
unique $\vecbs{B}{z}{G}$ based on the intersection of the various hemisphere and 
cone loci. The failure of two cones to intersect is precisely equivalent to a 
violation of the sine sum criterion, and hence an invalid specification of 
$\theta$ and $\phi$. The hemisphere parameter $h$ essentially decides which of 
the two cone intersections is used for $\vecbs{B}{z}{G}$.

\section{Conversions to Other Representations}
\seclabel{conversions}

Fused angles serve well in the analysis of body orientations, but even so, 
conversions to other representations are often required for mathematical 
computations such as rotation composition. The equations required for the 
conversion of the fused angles representation $F = (\psi,\theta,\phi,h) \in \F$ 
to and from tilt angles, rotation matrix and quaternion representations are 
presented in this section. Similar conversions are also provided for the tilt 
angles representation $T = (\psi,\gamma,\alpha) \in \T$. The proofs of the 
conversion equations are generally not difficult, but beyond the scope of this 
paper.

\subsubsection{Fused angles $\leftrightarrow$ Tilt angles}

The yaw parameters $\psi$ of these two representations are equal, so the
conversion from fused angles to tilt angles is completely summarised by
\begin{align}
\gamma &= \atantwo(\sin\theta,\sin\phi), \eqnlabel{fusedtogamma}\\
\alpha &= \acos\Bigl(h\sqrt{1-\sin^2\theta-\sin^2\phi}\Bigr), \eqnlabel{fusedtoalpha}
\end{align}
where for numerical computation one may use the identity
\begin{equation}
1-\sin^2\theta-\sin^2\phi \defeq \cos(\theta+\phi)\cos(\theta-\phi).
\end{equation}
We interestingly note from \eqnref{fusedtoalpha} that
\begin{equation}
\sin^2\theta + \sin^2\phi = \sin^2\alpha. \eqnlabel{sinesquaredsum}
\end{equation}
The conversion from tilt angles to fused angles is given by
\begin{align}
&\begin{aligned}
\theta &= \asin(\sin\alpha\sin\gamma), \\
\phi &= \asin(\sin\alpha\cos\gamma),
\end{aligned} &
\mspace{15mu} h &=
\begin{cases}
1& \text{if $\alpha \leq \hpi$,} \\
-1& \text{otherwise.}
\end{cases}
\eqnlabel{tilttofused}
\end{align}

\subsubsection{Tilt angles $\leftrightarrow$ Rotation matrix}

Based on the geometric definition of tilt angles given in 
\secref{tilt_angles_geometric}, the rotation matrix equivalent to
$T = (\psi,\gamma,\alpha) \in \T$ can be seen to be
\begin{equation}
R = 
\begin{bmatrix}
c_\gamma c_\beta + c_\alpha s_\gamma s_\beta & s_\gamma c_\beta - c_\alpha c_\gamma s_\beta & s_\alpha s_\beta \\
c_\gamma s_\beta - c_\alpha s_\gamma c_\beta & s_\gamma s_\beta + c_\alpha c_\gamma c_\beta & -s_\alpha c_\beta \\
-s_\alpha s_\gamma & s_\alpha c_\gamma & c_\alpha
\end{bmatrix}\!,
\eqnlabel{tilttorotmat}
\end{equation}
where $\beta = \psi + \gamma$, $s_x \defeq \sin{x}$ and $c_x \defeq \cos{x}$. 

The inverse conversion from a rotation matrix $R$ to the corresponding tilt 
angles representation $T$ is given by
\begin{align}
\gamma &= \atantwo(-R_{31},R_{32}), & \alpha &= \acos(R_{33}), \eqnlabel{rotmattogammaalpha}
\end{align}
where the fused yaw $\psi$ is calculated using 
\eqnrefs{rotmattopsicases}{rotmattopsitheta}, as presented below for the fused 
angles case.

\subsubsection{Fused angles $\leftrightarrow$ Rotation matrix}

The conversion of fused angles to a rotation matrix requires conversion via the 
tilt angles representation, using \eqnrefs{fusedtogamma}{fusedtoalpha}. 
Equation \eqnref{tilttorotmat} can then be used with slight simplification
as follows:
\begin{equation}
R = 
\begin{bmatrix}
c_\gamma c_\beta + c_\alpha s_\gamma s_\beta & s_\gamma c_\beta - c_\alpha c_\gamma s_\beta & s_\alpha s_\beta \\
c_\gamma s_\beta - c_\alpha s_\gamma c_\beta & s_\gamma s_\beta + c_\alpha c_\gamma c_\beta & -s_\alpha c_\beta \\
-s_\theta & s_\phi & c_\alpha
\end{bmatrix}\!.
\eqnlabel{fusedtorotmat}
\end{equation}
The conversion from $R$ back to $F$ follows from \eqnrefs{zGtotheta}{zGtohemi}.
If $R_m = \max\{R_{11},R_{22},R_{33}\}$ and $R_z = 1 - R_{11} - R_{22} + R_{33}$,
then the rotation matrix to fused angles conversion is most robustly given by
the following:
\pagebreak %
\begin{equation}
\tilde{\psi} =
\begin{cases}
\atantwo(R_{21}-R_{12}, 1+\tr(R))& \mspace{-2mu}\text{if $\tr(R) \geq 0$} \\
\atantwo(R_z, R_{21}-R_{12})& \mspace{-2mu}\text{if $R_m = R_{33}$} \\
\atantwo(R_{32}+R_{23}, R_{13}-R_{31})& \mspace{-2mu}\text{if $R_m = R_{22}$} \\
\atantwo(R_{13}+R_{31}, R_{32}-R_{23})& \mspace{-2mu}\text{if $R_m = R_{11}$}
\end{cases} \eqnlabel{rotmattopsicases}
\end{equation}\vspace{-1.0em}\begin{align}
\psi &= \wrap(2\tilde{\psi}), & \theta &= \asin(-R_{31}), \eqnlabel{rotmattopsitheta}\\
\phi &= \asin(R_{32}), & h &= \sign(R_{33}). \eqnlabel{rotmattophihemi}
\end{align}
Although it is possible to construct a much simpler expression for $\psi$ using 
\eqnref{psidefn}, this is not recommended due to the resulting numerical 
sensitivity near $\alpha = 0$.

\subsubsection{Tilt angles $\leftrightarrow$ Quaternion}

The conversion of a tilt angles rotation $T = (\psi,\gamma,\alpha) \in \T$ to 
the corresponding quaternion representation is robustly given by
\begin{equation}
q = (c_{\bar{\alpha}} c_{\bar{\psi}}, s_{\bar{\alpha}} c_{\bar{\psi}+\gamma}, s_{\bar{\alpha}} s_{\bar{\psi}+\gamma},
c_{\bar{\alpha}} s_{\bar{\psi}}), \eqnlabel{tilttoquat}
\end{equation}
where $\bar{\alpha} = \half\alpha$ and $\bar{\psi} = \half\psi$. In 
combination with \eqnref{quattopsitheta} for calculating the fused yaw $\psi$,
the inverse conversion from quaternion $q$ to tilt angles $T$ is given by
\begin{align}
\gamma &= \atantwo(wy-xz,wx+yz), \eqnlabel{quattogamma}\\
\alpha &= \acos\bigl(2(w^2 + z^2) - 1\bigr). \eqnlabel{quattoalpha}
\end{align}

\subsubsection{Fused angles $\leftrightarrow$ Quaternion}

The conversion from fused angles to quaternions is robustly given by
\begin{align}
q &= 
\begin{cases}
\dfrac{\tilde{q}_p}{\norm{\tilde{q}_p}}& \text{if $h = 1$}, \\[0.8em]
\dfrac{\tilde{q}_n}{\norm{\tilde{q}_n}}& \text{if $h = -1$},
\end{cases} \eqnlabel{fusedtoquat}\\[0.3em]
\tilde{q}_p &= \bigl( c_{\bar{\psi}} C_\alpha^+, s_\phi c_{\bar{\psi}} - s_\theta s_{\bar{\psi}},
s_\phi s_{\bar{\psi}} + s_\theta c_{\bar{\psi}}, s_{\bar{\psi}} C_\alpha^+ \bigr), \eqnlabel{fusedtoquatpos}\\
\tilde{q}_n &= \bigl( s_\alpha c_{\bar{\psi}}, c_{\bar{\psi}+\gamma} C_\alpha^-,
s_{\bar{\psi}+\gamma} C_\alpha^-, s_\alpha s_{\bar{\psi}} \bigr), \eqnlabel{fusedtoquatneg}
\end{align}
where $C_\alpha^+ = 1 + c_\alpha$ and $C_\alpha^- = 1 - c_\alpha$. The 
respective quaternion norms are analytically given by
\begin{align}
\norm{\tilde{q}_p} &= \sqrt{2 C_\alpha^+} = 2c_{\bar{\alpha}}, &
\norm{\tilde{q}_n} &= \sqrt{2 C_\alpha^-} = 2s_{\bar{\alpha}}. \eqnlabel{fusedtoquatnorms}
\end{align}
Note that $\alpha$ does not need to be computed in order to evaluate 
\eqnrefs{fusedtoquat}{fusedtoquatneg}, just $c_\alpha$ and $s_\alpha$. These can 
be obtained directly from \eqnref{fusedtoalpha} and \eqnref{sinesquaredsum}. 
Using \eqnrefs{rotmattopsitheta}{tilttoquat}, the fused angles representation of 
a quaternion $q = (w,x,y,z) \in \Q$ can be shown to be
\begin{align}
\psi &= \wrap\bigl(2\atantwo(z,w)\bigr), & \theta &= \asin\bigl(2(wy-xz)\bigr), \eqnlabel{quattopsitheta}\\
h &= \sign(w^2+z^2-\half), & \phi &= \asin\bigl(2(wx+yz)\bigr). \eqnlabel{quattophihemi}
\end{align}
Note that this expression for $\psi$ is insensitive to the quaternion magnitude, 
and far more direct than an expression derived from \eqnref{rotmattopsicases} 
would be. In fact, \eqnref{quattopsitheta} can conveniently be taken as the 
mathematical \emph{definition} of fused yaw. Note that the angle wrapping of 
$\psi$ is at most by a single multiple of $2\pi$.

\section{Singularity Analysis}
\seclabel{singularities}

When examining rotation representations, it is important to identify and
precisely quantify any singularities. Singularities are unavoidable in any
minimal parameterisation, and may occur in the form of:
\begin{enumerate}[label=(\roman*)]
 \item A rotation that cannot unambiguously be resolved into the required set of
       rotation parameters, \enumlabel{singularity_type_1}
 \item A rotation for which there is no equivalent parameterised representation
       that is unambiguous, \enumlabel{singularity_type_2}
 \item A rotation in the neighbourhood of which the sensitivity of the rotation
       to parameters map is unbounded. \enumlabel{singularity_type_3}
\end{enumerate}
The entries of a rotation matrix are a continuous function of the underlying 
rotation and lie in the interval $[-1,1]$. As such, from 
\eqnrefs{rotmattopsitheta}{rotmattophihemi} and the continuity of the 
appropriately domain-restricted arcsine function, it can be seen that the fused 
pitch and fused roll are continuous over the entire rotation space. Furthermore, 
the hemisphere parameter of the fused angles representation is uniquely and 
unambiguously defined over the rotation space. As a result, despite its discrete 
and thereby technically discontinuous nature, the hemisphere parameter is not 
considered to be the cause of any singularities in the fused angles 
representation.

The fused yaw parameter, on the other hand, can be seen from 
\eqnref{quattopsitheta} to have a singularity at $w = z = 0$, due to the 
singularity of $\atantwo$ at $(0,0)$. From \eqnref{tilttoquat}, this condition 
can be seen to be precisely equivalent to $\alpha = \pi$, the defining equation 
of the set of all rotations by 180\degree about axes in the $\vect{x}\vect{y}$ 
plane. The fused yaw singularity is a singularity of type 
\enumref{singularity_type_2} and \enumref{singularity_type_3} as per the 
characterisation given above, and corresponds to an essential discontinuity of 
the fused yaw map. Moreover, given any fused yaw singular rotation $R$, and any 
neighbourhood $U$ of $R$ in the rotation space $\text{SO}(3)$, for every
$\psi \in (-\pi,\pi]$ there exists a rotation in $U$ with a fused yaw of $\psi$. 
Conceptually, the fused yaw singularity can be seen as being as `far away' from 
the identity rotation as possible. This is by contrast not the case for Euler 
angles.

The tilt angles representation trivially has the same singularity in the fused 
yaw as the fused angles representation. In addition to this however, from 
\eqnref{rotmattogammaalpha}, the tilt axis angle $\gamma$ also has a singularity 
when $R_{31} = R_{32} = 0$. This corresponds to $\theta = \phi = 0$, or 
equivalently, $\alpha = 0$ or $\pi$---that is, either rotations of pure yaw, or 
rotations by 180\degree about axes in the $\vect{x}\vect{y}$ plane. The tilt 
angle parameter $\alpha$ is continuous by \eqnref{zGtoalpha} and the continuity 
of the arccosine function, and as such does not contribute any further 
singularities.

\section{Results and Properties of Fused Angles}
\seclabel{properties}

The fused angles representation possesses a remarkable number of subtle 
properties that turn out to be quite useful both mathematically and 
geometrically when working with them. One of these properties, relating to the 
axisymmetry of the representation, has already been stated without proof in 
\secref{review_eulerangles}. Other---more complex---properties of fused angles, 
involving for example the matching of fused yaws between coordinate frames, were 
invoked in \cite{Allgeuer2014} to derive a computationally efficient algorithm 
to calculate instantaneous measurements of the orientation of a body from sensor 
data. Some of the more basic but useful properties of fused angles are presented 
in this section.

\subsection{Fundamental Properties of Fused Angles}

The following fundamental properties of fused angles hold, and form a minimum
set of axiomatic conditions on the fused angles parameters.
\begin{itemize}
 \item A pure x-rotation by $\beta \in [-\hpi,\hpi]$ is given by the fused
       angles representation $(0,0,\beta,1) \in \F$.
 \item A pure y-rotation by $\beta \in [-\hpi,\hpi]$ is given by the fused
       angles representation $(0,\beta,0,1) \in \F$.
 \item A pure z-rotation by $\beta \in (-\pi,\pi]$ is given by the fused angles
       representation $(\beta,0,0,1) \in \F$.
 \item Applying a pure z-rotation to an arbitrary fused angles rotation is
       purely additive in fused yaw.
\end{itemize}
Further fundamental properties of fused angles include:
\begin{itemize}
 \item The parameter set $(\psi,\theta,\phi,h) \in \F$ is valid if and only if
       $\abs{\theta} + \abs{\phi} \leq \hpi$, i.e.\ the sine sum criterion is satisfied.
 \item The parameter set $(\psi,\theta,\phi,h) \in \F$ can be put into standard
       form by setting $h = 1$ if $\abs{\theta} + \abs{\phi} = \hpi$, and
       $\psi = 0$ if $\theta = \phi = 0$ and $h = -1$ (i.e.\ $\alpha = \pi$).
 \item Two fused angles rotations are equal if and only if their standard forms 
       are equal. Note that this identifies the $\alpha = \pi$ rotations due to 
       the fused yaw singularity there.
\end{itemize}

\subsection{Inverse of a Fused Angles Rotation}

The fused angles representation of the inverse of a rotation is intricately
linked to the fused angles parameters of a rotation. This is an almost
unexpected result when compared to, for example, Euler angles, but follows
trivially from the formulas and properties presented in this paper thus far.
Consider a fused angles rotation $(\psi,\theta,\phi,h)$ with an equivalent tilt
angles representation $(\psi,\gamma,\alpha)$. The parameters of the inverse
rotation are given by
\begin{align}
\psi_{inv} &= -\psi, \mspace{12mu} & \gamma_{inv} &= \wrap(\psi + \gamma - \pi), \eqnlabel{invpsigamma}\\
\alpha_{inv} &= \alpha, & \theta_{inv} &= \asin\bigl(-\sin\alpha\sin(\psi + \gamma)\bigr), \\
h_{inv} &= h, & \phi_{inv} &= \asin\bigl(-\sin\alpha\cos(\psi + \gamma)\bigr).
\end{align}
The leftmost equation in \eqnref{invpsigamma} represents a remarkable property of
fused yaw, one that other definitions of yaw such as ZYX Euler yaw do not
satisfy. This property is referred to as negation through rotation inversion. It
is worth noting that if a rotation has zero fused yaw, i.e.\ it is a pure tilt
rotation, the inverse fused pitch and roll also satisfy the negation through
rotation inversion property. That is,
\begin{equation}
\psi = 0 \;\iff\,
\begin{lbraced}
\psi_{inv} &= -\psi, & \theta_{inv} &= -\theta, \\
h_{inv} &= h, & \phi_{inv} &= -\phi.
\end{lbraced}
\end{equation}

\subsection{Characterisation of the Fused Yaw of a Quaternion}

For rotations away from the singularity $\alpha = \pi$, that is, for rotations
where the fused yaw is well-defined and unambiguous, inspection of
\eqnref{tilttoquat} reveals that the z-component of a quaternion
$q = (w,x,y,z) \in \Q$ is zero if and only if the fused yaw is zero. That is,
\begin{equation}
z = 0 \;\iff\; \psi = 0. \eqnlabel{zfusedyawzero}
\end{equation}
Furthermore, it can be seen that the quaternion corresponding to the fused yaw 
of the rotation can be constructed by zeroing the x and y-components of $q$ and 
renormalising. That is,
\begin{equation}
q_{yaw} = \tfrac{1}{\sqrt{w^2+z^2}}(w,0,0,z). \eqnlabel{quatyaw}
\end{equation}
This leads to one way of removing the fused yaw component of a 
quaternion---something that is a surprisingly common operation---using the 
expression
\begin{equation}
q_{tilt} = q_{yaw}\conj q = \tfrac{1}{\sqrt{w^2+z^2}}\Bigl(wq + z(z,y,-x,-w)\Bigr). \eqnlabel{quatnoyaw}
\end{equation}
The fused yaw can also be computed using \eqnref{quattopsitheta} and manually 
removed. Equations \eqnrefs{quatyaw}{quatnoyaw} fail only if $w = z = 0$, which 
is precisely equivalent to $\alpha = \pi$, the fused yaw singularity.

\subsection{Metrics over Fused Angles}

For the design of rotation space trajectories and other purposes, it is useful 
to be able to quantify the distance between two rotations using a metric. 
Assuming two fused angle rotations $F_1$ and $F_2$, and their corresponding tilt 
angles representations $T_1 = (\psi_1, \gamma_1, \alpha_1)$ and $T_2 = (\psi_2, 
\gamma_2, \alpha_2)$, two naturally arising metrics \cite{Huynh2009} are 
($\mspace{2mu}\cdot$ is the dot product):
\begin{align}
d_R(F_1,F_2) &= \norm{\log(R_1\trans R_2)}_F = 2\acos\bigl(\abs{q_1 \mspace{-3mu}\cdot q_2}\bigr) = \theta, \\
d_L(F_1,F_2) &= 1 - \cos\bigl(\mspace{-2mu}\tfrac{\theta}{2}\mspace{-1mu}\bigr) = 1 - \abs{q_1 \mspace{-3mu}\cdot q_2},
\end{align}
where $q_1$, $q_2$ are the corresponding quaternions, $\theta$ is the angle 
magnitude of the relative axis-angle rotation $(\vect{\hat{u}},\theta)$, and
\begin{equation}
q_1 \mspace{-3mu}\cdot q_2 = c_{\bar{\alpha}_1} c_{\bar{\alpha}_2} c_{\Delta\bar{\psi}}
+ s_{\bar{\alpha}_1} s_{\bar{\alpha}_2} c_{\Delta\bar{\psi}+\Delta\gamma},
\vspace{0.3em}
\end{equation}
where $\Delta\bar{\psi} = \half(\psi_1 - \psi_2)$,
$\,\Delta\gamma = \gamma_1 - \gamma_2$, $\bar{\alpha}_1 = \half\alpha_1$ and\linebreak %
$\bar{\alpha}_2 = \half\alpha_2$. Note that the Riemannian metric $d_R$ relates 
closely to spherical linear interpolation (slerp) \cite{Shoemake1985}, and as 
such serves as a metric of first choice. Actual computation of slerp is however 
still most efficiently done in the quaternion space. Direct interpolation of 
fused angles can give unexpected results in the general case, but for two 
rotations in the positive hemisphere it is a viable alternative, that for many 
applications will produce completely satisfactory results.

\section{Conclusions}
\seclabel{conclusions}

Two novel ways of parameterising a rotation were formally introduced in this 
paper. The main contribution of these, the fused angles representation, was 
developed to be able to describe a rotation in a way that yields insight into 
the components of the rotation in each of the three major planes of the 
Euclidean space. The second parameterisation that was introduced, tilt angles, 
was defined as an intermediate representation between fused angles and other 
existing representations. Nevertheless, the tilt angles representation proves to 
be geometrically, conceptually and mathematically useful. Many properties of the 
fused angles and tilt angles representations were derived, often in highlight of 
their simplicity, and the relations of these two representations to other 
commonly used representations were explicitly given. The computational 
efficiency of the two representations can be seen by inspection of our 
open-source implementation \cite{MatOctRotLibGithub}. Due to their many special 
properties, fused angles fill a niche in the area of rotation parameterisation 
that is left vacant by alternative constructs such as Euler angles and 
quaternions, and are expected to yield valuable information and results, in 
particular in applications that involve balance.

\bibliographystyle{IEEEtran}
\bibliography{IEEEabrv,ms}

\begin{thebibliography}{10}
\providecommand{\url}[1]{#1}
\csname url@samestyle\endcsname
\providecommand{\newblock}{\relax}
\providecommand{\bibinfo}[2]{#2}
\providecommand{\BIBentrySTDinterwordspacing}{\spaceskip=0pt\relax}
\providecommand{\BIBentryALTinterwordstretchfactor}{4}
\providecommand{\BIBentryALTinterwordspacing}{\spaceskip=\fontdimen2\font plus
\BIBentryALTinterwordstretchfactor\fontdimen3\font minus
  \fontdimen4\font\relax}
\providecommand{\BIBforeignlanguage}[2]{{%
\expandafter\ifx\csname l@#1\endcsname\relax
\typeout{** WARNING: IEEEtran.bst: No hyphenation pattern has been}%
\typeout{** loaded for the language `#1'. Using the pattern for}%
\typeout{** the default language instead.}%
\else
\language=\csname l@#1\endcsname
\fi
#2}}
\providecommand{\BIBdecl}{\relax}
\BIBdecl

\bibitem{Allgeuer2014}
P.~Allgeuer and S.~Behnke, ``Robust sensor fusion for biped robot attitude
  estimation,'' in \emph{Proceedings of 14th IEEE-RAS Int. Conference on
  Humanoid Robotics (Humanoids)}, Madrid, Spain, 2014.

\bibitem{Allgeuer2013a}
P.~Allgeuer, M.~Schwarz, J.~Pastrana, S.~Schueller, M.~Missura, and S.~Behnke,
  ``A {ROS}-based software framework for the {NimbRo-OP} humanoid open
  platform,'' in \emph{Proceedings of 8th Workshop on Humanoid Soccer Robots,
  IEEE-RAS Int. Conference on Humanoid Robots}, Atlanta, USA, 2013.

\bibitem{MatOctRotLibGithub}
\BIBentryALTinterwordspacing
P.~Allgeuer. (2014, Oct) {Matlab/Octave Rotations Library}. [Online].
  Available: \url{https://github.com/AIS-Bonn/matlab_octave_rotations_lib/}
\BIBentrySTDinterwordspacing

\bibitem{Palais2009}
B.~Palais, R.~Palais, and S.~Rodi, ``A disorienting look at {E}uler's theorem
  on the axis of a rotation,'' \emph{The American Mathematical Monthly}, pp.
  892--909, 2009.

\bibitem{Tomazic2011}
S.~Toma{\v z}i{\v c} and S.~Stan{\v c}in, ``Simultaneous orthogonal rotations
  angle,'' \emph{Electrotechnical Review}, no.~78, pp. 7--11, 2011.

\bibitem{Bauchau2003}
O.~Bauchau and L.~Trainelli, ``The vectorial parameterization of rotation,''
  \emph{Nonlinear Dynamics}, vol.~32, no.~1, pp. 71--92, 2003.

\bibitem{Trainelli2004}
L.~Trainelli and A.~Croce, ``A comprehensive view of rotation
  parameterization,'' in \emph{Proceedings of ECCOMAS}, 2004.

\bibitem{Argyris1982}
J.~Argyris, ``An excursion into large rotations,'' \emph{Computer Methods in
  Applied Mechanics and Engineering}, vol.~32, pp. 85--155, 1982.

\bibitem{Huynh2009}
D.~Huynh, ``Metrics for {3D} rotations: Comparison and analysis,'' \emph{J. of
  Math. Imaging and Vision}, vol.~35, no.~2, pp. 155--164, 2009.

\bibitem{Shoemake1985}
K.~Shoemake, ``Animating rotation with quaternion curves,'' in \emph{ACM
  SIGGRAPH computer graphics}, vol.~19, no.~3, 1985, pp. 245--254.

\end{thebibliography}

\end{document}